\documentclass[11pt,letterpaper]{article}
\usepackage[hyperref]{naaclhlt2018}
\usepackage{times}
\usepackage{amsmath, amsfonts}
\usepackage{latexsym}
\usepackage[utf8]{inputenc}
\usepackage{textcomp}
\usepackage{pgfplots}
\usepackage{enumitem}
\usetikzlibrary{positioning}
\pgfplotsset{width=7.5cm,compat=1.12}
\newlist{longenum}{enumerate}{5}
\setlist[longenum,1]{label=\roman*)}
\setlist[longenum,2]{label=\alph*)}
\setlist[longenum,3]{label=\arabic*)}
\setlist[longenum,4]{label=(\roman*)}
\setlist[longenum,5]{label=(\alph*)}

\usepackage{url}
\aclfinalcopy 
\def\aclpaperid{4} 

\title{Analysis of Bag-of-n-grams Representation's Properties \\
Based on Textual Reconstruction}

\author{Qi Huang\\
  New York University \\
  {\tt qh384@nyu.edu} \\\And
  Zhanghao Chen \\
  New York University \\
  {\tt zc807@nyu.edu} \\\And
  Zijie Lu \\
  New York University \\
  {\tt zl1298@nyu.edu} \\\And
  Ye Yuan \\
  New York University \\
  {\tt yy1650@nyu.edu} \\}

\date{}

\begin{document}
\maketitle
\begin{abstract}
Despite its simplicity, bag-of-n-grams sentence representation has been found to excel in some NLP tasks. However, it has not received much attention in recent years and further analysis on its properties is necessary. We propose a framework to investigate the amount and type of information captured in a general-purposed bag-of-n-grams sentence representation. We first use sentence reconstruction as a tool to obtain bag-of-n-grams representation that contains general information of the sentence. We then run prediction tasks (sentence length, word content, phrase content and word order) using the obtained representation to look into the specific type of information captured in the representation. Our analysis demonstrates that bag-of-n-grams representation does contain sentence structure level information. However, incorporating n-grams with higher order n empirically helps little with encoding more information in general, except for phrase content information.
\end{abstract}

\section{Introduction}
Though simple as it appears, bag-of-n-grams representation of textual data has been found to excel in many natural language processing (NLP) tasks, in particular sentiment analysis \cite{Cho:17}. This suggests that it may encode most information of a sentence. This paper aims to investigate the properties of a general-purposed bag-of-n-grams representation to reveal the amount and type of information it captures.

A good sentence representation paves the way for better performance in various NLP tasks, and various methods have been developed to generate a good sentence representation. Continuous-bag-of-words (CBOW) model \cite{Mikolov:13} is efficient to train, and performs well in many downstream tasks. However, it may have discarded the word order information and some semantics. Recently, neural network-based sentence representation models including Recursive Neural Networks (RecNN) \cite{Socher:12}, Convolutional Neural Networks (CNN) \cite{Kim:14} and Recurrent Neural Networks (RNN) \cite{Sutskever:14, Cho:14, Bahdanau:14} embeddings have shown advantages in generating general purpose sentence representation. They capture more syntactic and semantic structures of the sentences, but are computationally heavy.

On the other hand, a bag-of-n-grams embedding represents a sentence with a vector by summing over the n-gram embeddings, in theory richer in local order and syntactic information than CBOW and still computationally cheaper than neural networks based sentence representations. Its simplicity can come useful in certain situations, and a more thorough analysis of its properties is needed.

We propose a framework to perform detailed and systematic analysis of the properties of bag-of-n-grams sentence representations. To make our analysis more meaningful in a realistic sense, we analyze on general-purposed bag-of-n-grams representations. Sentence reconstruction gives us general-purposed embeddings, and we then test the obtained embeddings on prediction tasks including length, word content, phrase content, and word order prediction, each to reflect bag-of-n-grams' capacity of capturing a particular type of sentence information. We also report word level encoder-decoder model's performance on theses tasks as a baseline.

\section{Related Works}
Word-level distributed representations and their corresponding properties have been analyzed extensively. In contrast, study on bag-of-n-grams and its corresponding properties has been limited.

Studies such as \cite{Le:14} show that bag-of-n-grams model is usually considered deficient in dealing with data sparsity and poor generalization. In particular, Le shows that CBOW and bigram models perform poorly in encoding paragraph information, and bigram representation generally outperforms unigram. This induces the question whether this result can be generalized to sentence representation and whether a bigger n leads to even better performance. However, a more detailed and systematic analysis has not been done on the properties of bag-of-n-grams embeddings.

\textbf{Bag-of-n-grams Embedding}
Nonetheless, different embedding approaches of bag-of-n-grams have been proposed. For example, Li et al. \shortcite{Li:17} propose methods to train a distributed n-gram based representation in Neural Bag-of-Words, which vary in their loss functions’ design (context-guided based, text-guided based, and label-guided based). However, their work does not explore the impact of choices of n on model's performance.

In comparison, our work provides a more thorough analysis of bag-of-n-grams representation, and the methodology we adopt is more systematic and can be applied to other embedding analysis.

\textbf{Embedding Analysis Techniques}
Regarding analysis techniques, we are inspired by the following papers in the course of our research.

Adi et al. \shortcite{Adi:16} introduce prediction tasks to analyze if certain kinds of information are encoded in the vector representation of a sentence. They focus on distributed representations of sentence generated by CBOW and Encoder-Decoder model. In general, they provide a reasonable framework to analyze properties of distributed representations, and we decide to extend their analysis to bag-of-n-grams based sentence representation.

Arne Kohn \shortcite{Kohn:15} argues that evaluation of word embedding scheme on languages other than English is lacking, and the question whether word embeddings work the same way across languages has not been empirically evaluated. He concludes that the tested word embedding algorithms behave similarly across different languages. Following this line, we propose to also test our bag-of-n-grams sentence embedding on other languages, specifically languages which are morphologically complex such as Finnish and Turkish. We make this choice because previous computational linguistic research shows that language complexity can be measured by the language's morphological tier. As Juola \shortcite{Juola:98} shows, morphological complexity is a common measure for languages' overall complexity, and Turkish and Finnish's morphological complexity is higher than that of English. We expect the bag-of-n-grams based sentence reconstruction to suffer from a performance drop when embedding Finnish and Turkish sentence for this reason, and the result pattern across n should still be similar.

\section{Approach}
We aim to propose a framework to analyze the properties of general purpose bag-of-n-grams sentence representation, including the amount and type of information captured by the representation. In order to obtain general-purposed bag-of-n-grams representations, we use the sentence reconstruction as a tool to learn the representations. The intuition is that if a bag-of-n-grams sentence representation performs well in sentence reconstruction, it must contain most of the useful information in the original sentence. Therefore, we think it's a simple and logical way to use the bag-of-n-grams embeddings obtained from sentence reconstruction as the embedding for prediction tasks.

We then feed the bag-of-n-grams sentence representation as raw input to prediction tasks to further investigate what specific types of information and what amount of that information is encoded in the representation. We vary the choice of n from 1 to 5, and compare the results with those achieved by word-level RNN-encoder based sentence embeddings, which serve as the baseline. Finally, as an extra experiment we test our bag-of-n-grams representation for Finnish and Turkish, both of which are morphologically more complex languages.

\subsection*{Notation}
Let $S$ denote a sentence, and we use $w_i^j$ to represent the $j$-th $i$-gram of the sentence. There are in total $N_i$ $i$-grams of the sentence, and the $i$-gram representation of the sentence is $S_i =\{w_1^1, w_1^2, \cdots w_1^{N_1}, \cdots, w_i^1, w_i^2,\cdots w_i^{N_i}\}$. The $i$-gram $w_i^j$ is associated with a $K$-dimensional embedding $e_i^j$, and so the vector of bag-of-$i$-gram representation of a sentence will be $E_i = \sum_{j=1}^{N_i} e_i^j$. After summation, the bag-of-n-grams vector representation of a sentence $\bar{E}_i$ is given by
\begin{align*}
\bar{E}_n =  \sum_{i=1}^n E_i.
\end{align*}

\subsection{Sentence Reconstruction}
Inspired by RNN Encoder-Decoder model proposed by Cho et al. ~\shortcite{Cho:14}, we replace the encoder in our framework with a simple embedder that transforms a sentence $S$ to its bag-of-n-grams vector $\bar{E}_n$. We maintain the general structure of the decoder, which is an RNN that is trained to generate a sequence of words by predicting the next word $y_t$ given the hidden state $h_{\langle t\rangle}$ and its previous word $y_{t-1}$. The initial hidden state of the decoder $h_{\langle0\rangle}$ is the bag-of-n-grams vector output by the embedder, and the initial input $y_0$ is the starting of sentence (SOS) token. We then train the entire model end-to-end to reconstruct the original sentence from the embedding vector to achieve a general-purpose bag-of-n-grams sentence representation, which we feed as input for prediction tasks.

\subsection{Prediction Tasks}

\subsubsection*{Sentence Length}
With the bag-of-n-grams vector $\bar{E}_n \in \mathbb{R}^K$ of a sentence as input, we use a multi-layer perceptron (MLP) classifier to predict the length of the sentence. We formulate it as a multi-class classification, with several output classes according to preset length range. Grouping lengths into classes avoids unnecessarily too many classes while still maintaining the core goal of this task. A good performance on this task would suggest that the length of the original sentence is reasonably encoded into the bag-of-n-grams representation.

\subsubsection*{Word Content}
This task serves to determine whether the information of each individual word from the original sentence can still be identified after the summation over all n-grams from the original sentence that forms the resulted bag-of-n-grams sentence embedding. With the bag-of-n-grams vector $\bar{E}_n \in \mathbb{R}^K$ and the vector of a word $w \in \mathbb{R}^K$ directly extracted from the word's corresponding n-gram embedding as input, the MLP classifier's job is to determine whether the corresponding word is contained in the original sentence represented by $\bar{E}_n$. We formulate this task as a binary classification problem.

\subsubsection*{Phrase Content}
Similar to word content, this task serves to determine whether the information of a phrase from the original sentence can still be identified from the original sentence's bag-of-n-grams embedding. A fundamental difference between word-based representation and n-gram-based representation is how neighboring words, or phrases, are embedded. In theory, n-gram based representations embed a phrase as a singleton, while word-based representations do not. Therefore, for commonly seen phrase, n-gram based representation should be able to capture the information of a phrase more effectively. With the bag-of-n-grams embedding vector of a sentence $\bar{E}_n \in \mathbb{R}^K$ and the bag-of-n-grams embedding vector of a phrase $p \in \mathbb{R}^K$ as input, the MLP classifier's job is to determine whether the corresponding phrase is contained in the original sentence represented by $\bar{E}_n$.

\subsubsection*{Word Order}
This task evaluates how well the bag-of-n-grams representation can preserve the syntactical information of the order among words in the original sentence. We feed the bag-of-n-grams vector $\bar{E}_n \in \mathbb{R}^K$, and two word representations $e_1 \in \mathbb{R}^K, e_2 \in \mathbb{R}^K$, the classifier's job is to determine which word comes first in the sentence. We formulate this task as a binary classification as well.

\section{Experiment Settings}

\subsection{Dataset}
We perform experiments on the Tab-delimited Bilingual Sentence Pairs (English-French, English-Finnish) dataset \cite{Dataset}. The dataset is cleaned to remove repeated sentences and lowercased in order to avoid an issue of data sparsity. The dataset is then randomly shuffled and split into a training set (80\%) and a test set (20\%). We trim the dataset to have the first 20,000 pairs in training set for training, and first 5,000 pairs in test set for testing. SpaCy \cite{spacy2} is used for automatic tokenization. We keep 50,000 most frequent n-grams in a dictionary which is used to train our models to mitigate the sparsity issue. Any word not in the dictionary is mapped to a special token (UNK). We make sure that the dictionary contains a reasonable amount of higher order n-grams to avoid having higher order n-grams underrepresented.

\subsection{Sentence Reconstruction}
We train the proposed bag-of-n-grams based model. We also train an RNN autoencoder, which serves as a baseline.

All the encoder and decoder networks are Gated Recurrent Units (GRU) networks \cite{Cho:14} with 256 hidden units each. In all cases, we use a multilayer network with a softmax activation layer to compute the conditional probability of each target word.

We use a stochastic gradient descent (SGD) algorithm to train each model. Random teacher forcing enabled with a probability of 0.5 is used to make the models converge faster. Each model is trained for 20 epochs with a learning rate of 0.01.

To measure the closeness of our reconstructed sentences and the original sentences from where n-grams are drawn, we adopt BLEU-clip \cite{Papineni:02} as our metric.

\subsection{Prediction Tasks}
A simple MLP model with two hidden layers of size 128 and 64 respectively is used for classification in the length, word content, phrase content and word order tasks. Sentence representations obtained from sentence reconstruction task are directly fed as raw input to the MLP. The representations are kept fixed during the training. After hyperparameter tuning, we use a learning rate of 0.001 for all the prediction tasks. Their performances are reported as the percentage prediction accuracy.

For phrase content prediction, since the number of bigrams and trigrams are the largest in the clipped dictionary, we choose to only use them (phrases of length 2 and 3) as phrase content task input.

\section{Results}
In this section we provide a detailed description of our experimental results along with their analysis.

\begin{figure}[h]
\begin{tikzpicture}
\begin{axis}[
    ylabel={BLEU-Clip score},
    symbolic x coords={RNN, 1, 2, 3, 4, 5},
    ymin = 0.2, ymax = 1,
    height=0.35\textwidth,
    ytick={0,0.2,0.4,0.6,0.8,1.0},
    legend pos=north east,
    ymajorgrids=true,
    grid style=dashed,
]

\addplot[
    color=blue,
    mark=square,
    ]
    coordinates {
    (RNN, 0.685)(1,0.682)(2, 0.561)(3, 0.582)(4, 0.568)(5, 0.571)
    };
    
\addplot[
    color=orange,
    mark=square,
    ]
    coordinates {
    (RNN, 0.504)(1,0.45)(2, 0.35)(3, 0.353)(4, 0.336)(5, 0.320)
    };
    
\addplot[
    color=red,
    mark=square,
    ]
    coordinates {
    (RNN, 0.606)(1,0.58)(2, 0.469)(3, 0.482)(4, 0.467)(5, 0.461)
    };
    \legend{Short sentences, Long sentences, Overall}
    
\end{axis}
\end{tikzpicture}
\caption{Sentence Reconstruction Score}
\end{figure}
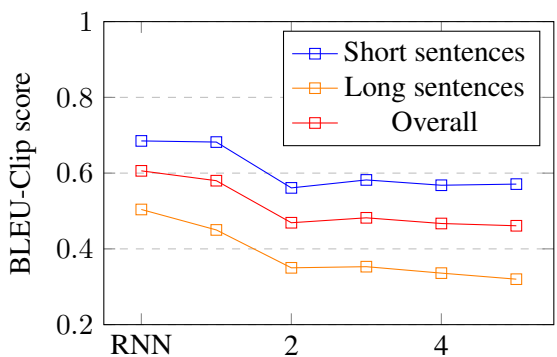
\subsection{Sentence Reconstruction}
We use sentence reconstruction on English to train our bag-of-n-grams sentence representations, each with n from 1 to 5 as well as a RNN encoder. Sentence reconstruction's BLEU-clip score could be used as a rough indicator of the overall amount of information encoded in the representation. As Figure 1 shows, the performance of unigram rivals that of the encoder with BLEU-clip scores of around 0.6 in English, while all other choices of n yield similar results that are worse. More specifically, performance drops as n increases. There is a 0.1 to 0.2 gap between the score on short sentences (length $\leq$ 6) and on long sentences for all models, indicating that our bag-of-n-grams representation suffers when encoding longer sentences to a similar extent as the baseline RNN Encoder. This suggests that when embedding longer sentences, bag-of-n-grams representation with bigger n may not offer more information, and may actually add more noise. We examine this more in the following prediction tasks.

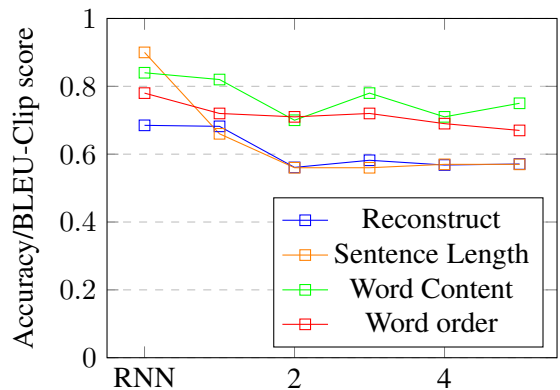
\begin{figure}[h]
\begin{tikzpicture}
\begin{axis}[
    ylabel={Accuracy/BLEU-Clip score},
    symbolic x coords={RNN, 1, 2, 3, 4, 5},
    ymin = 0, ymax = 1,
    height=0.38\textwidth,
    ytick={0,0.2,0.4,0.6,0.8,1.0},
    legend pos=south east,
    ymajorgrids=true,
    grid style=dashed,
]

\addplot[
    color=blue,
    mark=square,
    ]
    coordinates {
    (RNN, 0.685)(1,0.682)(2, 0.561)(3, 0.582)(4, 0.568)(5, 0.571)
    };
    
\addplot[
    color=orange,
    mark=square,
    ]
    coordinates {
    (RNN, 0.9)(1,0.66)(2, 0.56)(3, 0.56)(4, 0.57)(5, 0.57)
    };
    
\addplot[
    color=green,
    mark=square,
    ]
    coordinates {
    (RNN, 0.84)(1,0.82)(2, 0.7)(3, 0.78)(4, 0.71)(5, 0.75)
    };
   
\addplot[
    color=red,
    mark=square,
    ]
    coordinates {
    (RNN, 0.78)(1,0.72)(2, 0.71)(3, 0.72)(4, 0.69)(5, 0.67)
    };
    \legend{Reconstruct, Sentence Length, Word Content, Word order}
    
\end{axis}
\end{tikzpicture}
\caption{Partial prediction tasks results}
\end{figure}

\subsection{Prediction Task Results}
\subsubsection*{Sentence Length}
As Figure 2 shows, RNN-encoder has by far the best performance in sentence length task, correctly predicting 90\% of the test samples; unigram follows at 66\% while other bag-of-n-grams with n from 2 to 5 exhibit an accuracy at around 57\%. 

One explanation on why any sentence representation may capture sentence length information is that, as indicated by \cite{Adi:16}, the norm of the representation vector plays an important role in encoding sentence length. We perform a similar experiment on bag-of-n-grams based embeddings. As shown in Figure 3, in general, the norm of the bag-of-n-grams based embedding vectors increases as the sentence length increases. Furthermore, RNN-encoder based embedding vectors display a similar trend. This is quite remarkable, as RNN-encoder's result reinforces our proposition about bag-of-n-grams representation, and it further suggests that both models could possibly embed sentence length in the vectors' norm. However, the overall accuracy actually drops as n increases, and remains relatively constant across n higher than 1. Except for the remarkable performance of unigram, the choice of n does not seem to affect the amount of length information encoded in bag-of-n-grams representation. This indicates that, at least in encoding sentence length information, higher order n may not help.

\begin{figure}[h]
\begin{tikzpicture}
\begin{axis}[
	xlabel={Sentence Length},
    ylabel={Vector Norm},
    symbolic x coords={1-2, 3-4, 5-6, 7-8, 9-10, 11-12, 13-14},
    ymin = 10, ymax = 200,
    ytick={20, 50,100, 150, 200},
    legend pos=north west,
    ymajorgrids=true,
    grid style=dashed,
]

\addplot[
    color=blue,
    mark=square,
    ]
    coordinates {
    (1-2, 35.83)(3-4, 45.52)(5-6, 53.55)(7-8, 61.65)(9-10, 78.18)(11-12, 78.18)(13-14, 85.13)
    };
    
\addplot[
    color=orange,
    mark=square,
    ]
    coordinates {
    (1-2, 40.82)(3-4,53.41)(5-6, 64.18)(7-8, 74.52)(9-10, 84.1)(11-12, 93.57)(13-14, 102.04)
    };
    
\addplot[
    color=green,
    mark=square,
    ]
    coordinates {
    (1-2, 33.69)(3-4,49.87)(5-6, 63.57)(7-8, 105.25)(9-10, 85.59)(11-12, 95.51)(13-14, 102.49)
    };
   
\addplot[
    color=red,
    mark=square,
    ]
    coordinates {
    (1-2, 32.91)(3-4,49.94)(5-6, 64.42)(7-8, 76.96)(9-10, 135.66)(11-12, 97.49)(13-14, 104.12)
    };
    
\addplot[
    color=olive,
    mark=square,
    ]
    coordinates {
    (1-2, 32.78)(3-4,49.68)(5-6, 82.89)(7-8, 77.41)(9-10, 81.04)(11-12, 98.29)(13-14, 104.25)
    };
    
\addplot[
    color=purple,
    mark=square,
    ]
    coordinates {
    (1-2, 18.77)(3-4, 19.9)(5-6, 20.85)(7-8, 23.2)(9-10, 27.4)(11-12, 28.1)(13-14, 35.57)
    };
    \legend{unigram, bigram, 3-gram, 4-gram, 5-gram, RNN}
    
\end{axis}
\end{tikzpicture}
\caption{Representation vector norm}
\end{figure}
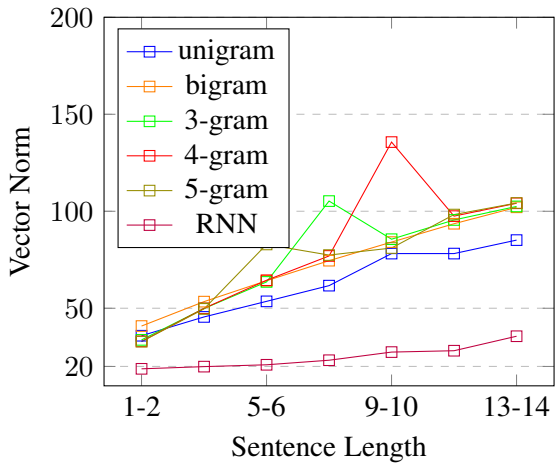

\subsubsection*{Word Content}
Unigram model has the best word content result with accuracy at over 80\%, while other models yield accuracy from 70\% to 78\% with no obvious correlation with n. We then divide the words for prediction into five categories by their frequency of appearance. We use vocabulary index to represent the frequency of the word (i.e. word with index 1 is the most frequent word).  By evaluating the prediction accuracy of each model on these groups of words separately, we study whether bag-of-n-grams representation's word content performance with each choice of n is related to word frequency. We observe that for each choice of n, the model is able to predict the occurrence of the most frequent words accurately. In addition, the model of each n struggles to predict the occurrence of words in the range of [500, 1000). An interesting phenomenon is that the ability to predict the presence of unknown words deteriorates as n increases, presumably due to vocabulary clipping.

\begin{table}[h]
\begin{center}
\begin{tabular}{ |c|c|c|c|c|c| } 
\hline
vocab index& 1 & 2 & 3 & 4 & 5 \\
\hline
[0, 100) &0.97&0.74&0.97&0.93&0.92\\
\hline
[100,500)&0.71&0.54&0.74&0.65&0.62\\ 
\hline
[500, 1000)&0.60&0.56&0.51&0.55&0.51\\
\hline
[1000, 6431)&0.71&0.75&0.56&0.74&0.66\\
\hline
unknown&0.96&0.81&0.26&0.16&0.06\\
\hline
\end{tabular}
\caption{Word Content accuracy with respect to word frequency}
\end{center}
\end{table}

\subsubsection*{Phrase Content}
We notice that the accuracy of predicting the presence of a phrase increases as n increases from 1 to 4. There is a decrease in the performance when using bag-of-5-grams, probably due to vocabulary clipping. However, bag-of-5-grams still outperforms bag-of-unigram and bag-of-bigram representation. Bag-of-unigram representation turns out not to be able to predict whether a two-word phrase occurs in the sentence. This indicates that bag-of-n-grams representation with bigger n does encode more information of a phrase's presence. This may be caused by the fact that bag-of-n-grams treats an n-gram ($n\geq 2$) as a singleton instead of treating it as a combination of n words. The co-occurrence of a bigram and those n-grams containing it seems to help inscribe its occurrence. This can also be applied to 3-gram as observed.

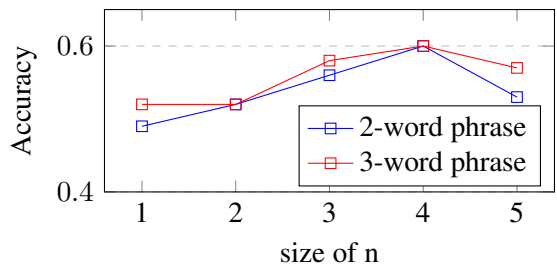
\begin{figure}[h]
\begin{tikzpicture}
\begin{axis}[
	xlabel={size of n},
    ylabel={Accuracy},
    height=0.25\textwidth,
    symbolic x coords={1, 2, 3, 4, 5},
    ymin = 0.4, ymax = 0.65,
    ytick={0,0.2,0.4,0.6,0.8,1.0},
    legend pos=south east,
    ymajorgrids=true,
    grid style=dashed,
]

\addplot[
    color=blue,
    mark=square,
    ]
    coordinates {
    (1,0.49)(2, 0.52)(3, 0.56)(4, 0.6)(5, 0.53)
    };
    
\addplot[
    color=red,
    mark=square,
    ]
    coordinates {
    (1,0.52)(2, 0.52)(3, 0.58)(4, 0.6)(5, 0.57)
    };
    \legend{2-word phrase, 3-word phrase}
\end{axis}
\end{tikzpicture}
\caption{Phrase Content Result}
\end{figure}

\subsubsection*{Word Order}
RNN-encoder achieves an accuracy at 78\%, outperforming all bag-of-n-grams models. To investigate how the distance between two words affects the prediction, we divide the word pairs into 5 categories, where the distance is the number of words between the two words. We notice that when the distance is smaller than 4, the prediction accuracy generally increases as distance increases. We also notice within the same distance category, the prediction accuracy does not increase as n increases. This suggests that when treated as a singleton, an n-gram ($n\geq 2$) does not encode extra information of the orders of the words contained.

\begin{table}[h]
\begin{center}
\begin{tabular}{ |c|c|c|c|c|c| } 
\hline
distance& 1 & 2 & 3 & 4 & 5 \\
\hline
0 &0.73&0.72&0.73&0.70&0.70\\
\hline
1&0.75&0.75&0.76&0.73&0.74\\ 
\hline
2&0.79&0.78&0.77&0.76&0.77\\
\hline
3&0.83&0.80&0.80&0.75&0.78\\
\hline
$\geq4$&0.70&0.78&0.74&0.75&0.72\\
\hline
\end{tabular}
\caption{Word order accuracy with respect to word distance}
\end{center}
\end{table}

\subsection{Bag-of-n-gram-based representation of Morphologically Complex Languages}
As is shown in Figure 5, all models have poor BLEU-clip score in sentence reconstruction on Finnish and Turkish with the score hovering around 0.40, while all models achieve above 0.46 on English. This meets with our expectation, as both Finnish and Turkish have higher morphological complexity than English. However, we observe the similar pattern in sentence reconstruction of all three languages that RNN encoder and unigram model outperform all other higher-order bag-of-n-grams models, whose performance stays the same or slightly drops. This observation reinforces our previous conclusion that higher-order n-grams may not offer more useful information of the sentence and may actually add more noise. It further suggests that our findings on the properties of n-gram have the potential to transfer across languages of different morphological complexity.

\begin{figure}[h]
\begin{tikzpicture}
\begin{axis}[
    ylabel={BLEU-Clip score},
    symbolic x coords={RNN, 1, 2, 3, 4, 5},
    ymin = 0.3, ymax = 0.8,
    height=0.35\textwidth,
    ytick={0,0.2,0.4,0.6,0.8,1.0},
    legend pos=north east,
    ymajorgrids=true,
    grid style=dashed,
]

\addplot[
    color=blue,
    mark=square,
    ]
    coordinates {
    (RNN, 0.685)(1,0.682)(2, 0.561)(3, 0.582)(4, 0.568)(5, 0.571)
    };

\addplot[
    color=green,
    mark=square,
    ]
    coordinates {
    (RNN, 0.408)(1,0.419)(2, 0.385)(3, 0.379)(4, 0.375)(5, 0.375)
    };

\addplot[
    color=red,
    mark=square,
    ]
    coordinates {
    (RNN, 0.448)(1,0.408)(2, 0.374)(3, 0.364)(4, 0.362)(5, 0.362)
    };
    
    \legend{English, Finnish, Turkish}
    
\end{axis}
\end{tikzpicture}
\caption{Reconstruction Scores of English, Finnish and Turkish}
\end{figure}
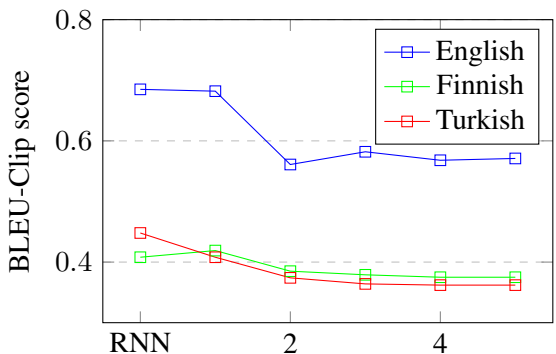

\section{Conclusion and Future Work}
We present a systematic set of experiments to perform an analysis of bag-of-n-grams sentence representation, specifically to answer the question of what kind of information it contains and how it may vary as n varies. Our results lead to the following conclusions.

\begin{itemize}
\setlength\itemsep{0.25em}
    \item Bag-of-n-grams sentence representation, which is capable of encoding general sentence information, contains a non-trivial amount of sentence length, word presence, phrase presence and word order information.
    \item General purpose bag-of-n-grams representation with higher order n does not necessarily encode more useful information. Unigram outperforms all other choices of n on nearly all tasks, while higher n's performance stays relatively the same or even decrease except in phrase content task.
    \item Phrase occurrence is better encoded in bag-of-n-grams representation with higher order n. This also suggests that when treated as a singleton, phrase information doesn't correlate with other structure level information such as word order and word content.
    \item Finally, though reconstruction score drops overall, the pattern observed above is still similar in our extra experiment on morphologically more complex languages, i.e. Finnish and Turkish. This further reinforces that the above conclusion holds across languages of different levels of complexity.
\end{itemize}

In our research, there are definitely interesting phenomena that await future exploration, and some aspects of our experiment could be improved. We could design a synthetic dataset that better accounts for sparsity issue, or incorporating attention mechanism in the training/analysis section to obtain a more insightful result. Due to the scope limit to this project, we decide to leave these as future work.

\subsection*{Collaboration Statement}
All four members collaborate on writing the paper. Qi Huang mainly works on Abstract, Introduction, Related Works, Approach, Results and Conclusion. Zhanghao Chen mainly works on Introduction, Approach, Experiment Settings, Results and running experiments. Zijie Lu mainly works on Introduction, Related Works, Approach, Results and running experiments. Ye Yuan mainly works on Approach and Results.

\subsection*{Public Codebase}
The codebase of this project is publicly available at
\href{https://github.com/HQ01/BOWMIAN}{https://github.com/HQ01/BOWMIAN}.

\bibliography{naaclhlt2018}
\bibliographystyle{acl_natbib}

\appendix

\end{document}